\documentclass[runningheads]{llncs}
\usepackage[T1]{fontenc}
\usepackage{graphicx}
\usepackage{amsmath}
\usepackage{amssymb}
\usepackage{multirow}
\usepackage{booktabs}
\usepackage{adjustbox}
\usepackage{cleveref}
\usepackage{tabularx}

\newcolumntype{C}{>{\centering\arraybackslash}X}

\begin{document}
\begin{sloppypar}
\title{RET-CLIP: A Retinal Image Foundation Model Pre-trained with Clinical Diagnostic Reports}
%
%
\author{Jiawei Du\inst{1} \and 
Jia Guo\inst{1} 
\and Weihang Zhang\inst{1} \and
Shengzhu Yang\inst{1} \and
Hanruo Liu\inst{1,2,3} \and
Huiqi Li\textsuperscript{1}\thanks{Corresponding author.} \and
Ningli Wang\inst{4}}
%
\authorrunning{Jiawei Du et al.}
%
\institute{Beijing Institute of Technology, Beijing, China \and
Beijing Tongren Hospital, Beijing, China \and
Capital Medical University, Beijing, China \and
Beijing Institute of Ophthalmology, Beijing, China\\
\email{\{jiaweidu, huiqili\}@bit.edu.cn}}
\maketitle              
\begin{abstract}
The Vision-Language Foundation model is increasingly investigated in the fields of computer vision and natural language processing, yet its exploration in ophthalmology and broader medical applications remains limited. The challenge is the lack of labeled data for the training of foundation model. To handle this issue, a CLIP-style retinal image foundation model is developed in this paper. Our foundation model, RET-CLIP, is specifically trained on a dataset of 193,865 patients to extract general features of color fundus photographs (CFPs), employing a tripartite optimization strategy to focus on left eye, right eye, and patient level to reflect real-world clinical scenarios. Extensive experiments demonstrate that RET-CLIP outperforms existing benchmarks across eight diverse datasets spanning four critical diagnostic categories: diabetic retinopathy, glaucoma, multiple disease diagnosis, and multi-label classification of multiple diseases, which demonstrate the performance and generality of our foundation model. The sourse code and pre-trained model are available at https://github.com/sStonemason/RET-CLIP.

\keywords{Vision-Language Pre-training \and Foundation Model \and Retinal Fundus Image.}
\end{abstract}
\section{Introduction}
Foundation models trained on large-scale, multi-task datasets are now becoming increasingly popular and have achieved success in the fields of computer vision and natural language processing. Foundation models excel in generalization in feature extraction, offering significant potential for addressing the complex challenges of clinical applications. However, the development of medical foundation models is still in its nascent phase, primarily hindered by the lack of high-quality data and concerns around patient privacy. Although initial efforts have been made \cite{zhangContrastiveLearningMedical2022,wang2022medclip,zhong2023chatradio,lin2023pmc,moor2023med,eslami2023pubmedclip,moor2023foundation}, the effectiveness of these models, particularly in analyzing retina fundus images, has yet to meet expectations, underscoring the urgent need for focused advancements in this area.

In the clinical diagnosis and treatment of ocular diseases, medical imaging, such as color fundus photography (CFP), and the detailed image interpretations and diagnostic reports written by professional ophthalmologists are indispensable. This makes the clinics of ophthalmology inherently rich in image-text multi-modality data, which holds significant potential for enhancing clinical applications. RETFound \cite{zhou2023foundation} is a foundation model for retinal images based on self-supervised learning. However, it solely utilizes image data and overlooks the equally vast amount of clinical diagnostic text. To address this limitation, CLIP \cite{radford2021learning}, a powerful vision-language self-supervised paradigm, is widely explored in foundation models. By aligning the information of image and text in a shared representation space using a large corpus of image-text pairs, CLIP-style models can understand and associate visual content with natural language information. This results in feature representations with stronger generalization capabilities. Many studies focus on training vision-text models in the medical field \cite{zhangContrastiveLearningMedical2022,wang2022medclip,lin2023pmc,zhang2023pmc,silva2023foundation,huang2021gloria,wu2023medklip,baliah2023exploring}. PMC-CLIP \cite{lin2023pmc} collects image-description pairs from large amount of scientific documents and trains a CLIP-style model based on them. FLAIR \cite{silva2023foundation} is a pre-trained vision-language model designed to understand retinal fundus images. The textual data utilized in such research often comes from captions in medical papers or through the manual annotation of simple labels. However, clinical diagnostic reports, rich in valuable textual information, remain underutilized in this context.

Moreover, the conventional approaches often involve treating CFPs of individual eyes as separate entities during model training. This necessitates the extraction of information corresponding to each eye from the original clinical diagnostic reports, which may not always clearly differentiate between left and right eyes. The manual processing involved in this procedure requires specialized knowledge and could introduce errors and increase costs significantly due to the potential for human-induced noise. Conversely, considering both eyes of a patient together provides a more holistic and clinically meaningful approach in clinical scenarios.

To alleviate the above issues, we have the following contributions in this paper: Firstly, we propose a vision-language foundation model for CFPs, named RET-CLIP, which we believe is the first attempt to leverage clinical diagnostic reports to build a retinal foundation model, enriching the model's visual encoding capabilities with practicality and authenticity. The diagnostic reports in Chinese are included, extending the linguistic versatility of the research domain beyond English. Secondly, a novel strategy is proposed to decouple the information of left and right eyes in diagnostic reports, which is a simple yet effective paradigm for building a retinal foundation model. In practical scenarios, diagnostic reports are usually patient-level, mixing information from both eyes, which brings a big challenge for directly using CLIP to build foundation models. The proposed monocular and patient-level contrastive learning approach can handle this challenge in the ophthalmology domain. Lastly, our model achieves state-of-the-art performance across diverse tasks and datasets, confirming the effectiveness of the proposed training strategy.


\section{Method}
\subsection{Data Collection and Preprocessing}
\textbf{Dataset acquisition.}
We collected a dataset of retina fundus binocular images-text triplets (RET-Clinical) at the patient level for RET-CLIP. The dataset includes a total of 193,865 samples from Beijing Tongren Hospital, Beijing, China. Each patient's triplet includes two CFPs for left and right eyes, alongside a clinical diagnostic report. 

\textbf{Data preprocessing and augmentation.}
For the CFPs, all of them are resized to \(512\times512\). The augmentation includes random crop followed by resizing to \(224\times224\), random horizontal flipping, color jitter, and image normalization. For diagnostic reports, we focus on correcting typos and consecutive punctuation errors caused by human input, restoring abbreviations to their full expressions, unifying mixed Chinese and English expressions into Chinese to align with our text encoder's language capabilities, and ensuring the text is coherent and grammatically correct by manual scrutiny. It's important to highlight that the preprocessing of text data only involves basic text standardization mentioned above, avoiding the need for advanced clinical knowledge or modifications that may alter the original content or meaning. 

\subsection{Model Architecture}
\begin{figure}
\includegraphics[width=\textwidth]{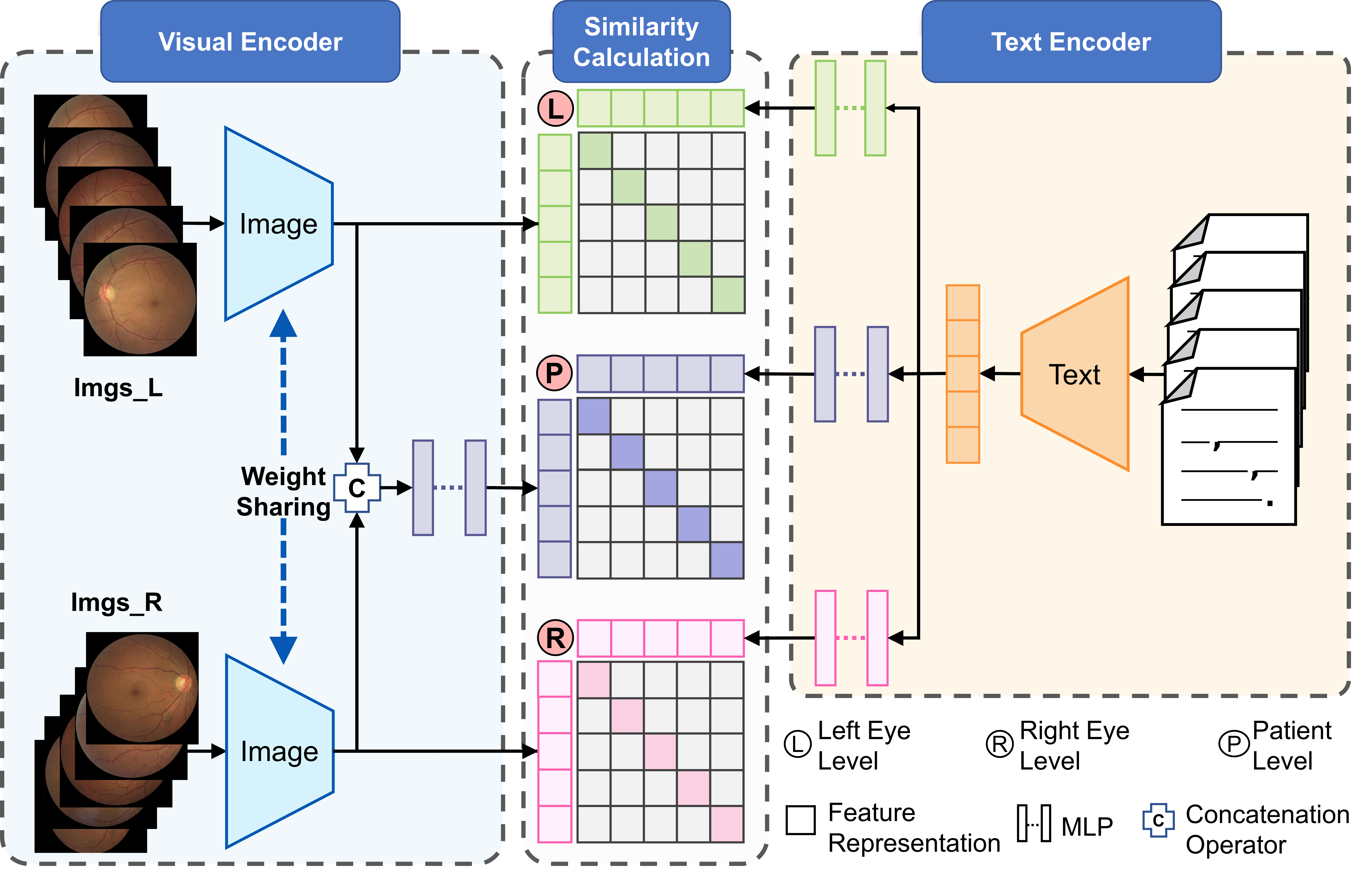}
\caption{Overview of the RET-CLIP foundation model.} \label{fig1}
\end{figure}

As shown in \Cref{fig1}, we trained a Visual-Language model called RET-CLIP under the CLIP paradigm using our constructed binocular images-text triplets. RET-CLIP consists of a visual encoder and a text encoder, which extract image features from CFPs and text features from clinical diagnostic reports, respectively. During pre-training, image-text contrastive learning is performed at the monocular and patient level jointly. Patient level examines data features from a holistic patient perspective, effectively leveraging the information in raw data while minimizing the interference of manual preprocessing in the pre-training phase. Concurrently, the binocular level guides the model towards acquiring finer-grained features than the patient level. Combined together, these methodologies can improve RET-CLIP's performance.

Given a mini-batch containing $N$ binocular images-text triplets (i.e., \(N\) patients), \(\mathcal{D}=\{(\mathcal{I}_{1}^{l},\mathcal{I}_{1}^{r},\mathcal{T}_{1}),\cdots,(\mathcal{I}_{N}^{l},\mathcal{I}_{N}^{r},\mathcal{T}_{N})\}\), where \(\mathcal{I}_{i}^{l}\), \(\mathcal{I}_{i}^{r}\) and  \(\mathcal{T}_{i}\) represents the CFP of left eye, the CFP of right eye and the diagnostic report of the $i$th patient, respectively. The visual encoder takes \(\mathcal{I}_{i}^{l}\) and \(\mathcal{I}_{i}^{r}\) as input, while the text encoder is fed with \(\mathcal{T}_{i}\).

\textbf{Visual encoder.}
The left and right \((\mathcal{I}^{l},\mathcal{I}^{r})\) CFPs for a patient are encoded to the embedding dimension of \(d\) using a ViT-based \cite{dosovitskiy2020image} encoder \(\Phi_{v}(\cdot)\) respectively:
\begin{equation}
\boldsymbol{V}^{l}=\Phi_{v}(\mathcal{I}^{l}) \in\mathbb{R}^{d},\boldsymbol{V}^{r}=\Phi_{v}(\mathcal{I}^{r}) \in\mathbb{R}^{d}.
\label{eq1}
\end{equation}
where \(\boldsymbol{V}^{l}\) and \(\boldsymbol{V}^{r}\) represent the image features of the left and right eye, respectively. Next, concatenation and a simple Multilayer Perceptron (MLP) \(F_{v}(\cdot)\) are employed to merge the image features of left and right eyes to derive comprehensive patient-level image features:
\begin{equation}
    \boldsymbol{V}^{p}=F_{v}(\boldsymbol{V}^{l} \oplus \boldsymbol{V}^{r}) \in\mathbb{R}^{d},
    \label{eq2}
\end{equation}
where \(\oplus\) denotes concatenation.

\textbf{Text encoder.}
For a given patient's diagnostic report \(\mathcal{T}\), a BERT-based \cite{devlin2018bert} encoder \(\Phi_{t}(\cdot)\) is implemented to encode the clinical descriptions with a text token of length \(l\):
\begin{equation}
    \boldsymbol{T}=\Phi_{t}(\mathcal{T}) \in\mathbb{R}^{l\times d}, \mathrm{~}\boldsymbol{T}_{0}\in\mathbb{R}^{d},
    \label{eq4}
\end{equation}
where \(\boldsymbol{T}\) denotes the sentence embedding, \(\boldsymbol{T}_{0}\) denotes the embedding for [CLS] token. We then implement three stacked two-layer nonlinear MLPs \(F_{l}(\cdot)\), \(F_{r}(\cdot)\), \(F_{p}(\cdot)\) to decouple \(\boldsymbol{T}_{0}\) into textual features representing the left eye, right eye, and patient level, termed as \(\boldsymbol{T}^{l}\), \(\boldsymbol{T}^{r}\), and \(\boldsymbol{T}^{p}\), respectively:
\begin{equation}
    \boldsymbol{T}^{l}=F_{l}(\boldsymbol{T}_{0}),\mathrm{~}\boldsymbol{T}^{r}=F_{r}(\boldsymbol{T}_{0}),\mathrm{~}\boldsymbol{T}^{p}=F_{p}(\boldsymbol{T}_{0}), \mathrm{~}\boldsymbol{T}^{l},\boldsymbol{T}^{r},\boldsymbol{T}^{p}\in\mathbb{R}^{d}.
    \label{eq5}
\end{equation}

\subsection{Training Objective}
For the provided mini-batch, termed as \(\mathcal{D}\), the extracted feature set \(\mathcal{F}\), which is \(\{(\boldsymbol{V}_{1}^{l},\boldsymbol{V}_{1}^{r},\boldsymbol{V}_{1}^{p},\boldsymbol{T}_{1}^{l},\boldsymbol{T}_{1}^{r},\boldsymbol{T}_{1}^{p}),\cdots,(\boldsymbol{V}_{N}^{l},\boldsymbol{V}_{N}^{r},\boldsymbol{V}_{N}^{p},\boldsymbol{T}_{N}^{l},\boldsymbol{T}_{N}^{r},\boldsymbol{T}_{N}^{p})\}\), is then divided into three subsets: \(\mathcal{F}^{l}=\{(\boldsymbol{V}_{1}^{l},\boldsymbol{T}_{1}^{l}),\cdots,(\boldsymbol{V}_{N}^{l},\boldsymbol{T}_{N}^{l})\}\), \(\mathcal{F}^{r}=\{(\boldsymbol{V}_{1}^{r},\boldsymbol{T}_{1}^{r}),\cdots,(\boldsymbol{V}_{N}^{r},\boldsymbol{T}_{N}^{r})\}\), and \(\mathcal{F}^{p}=\{(\boldsymbol{V}_{1}^{p},\boldsymbol{T}_{1}^{p}),\cdots,(\boldsymbol{V}_{N}^{p},\boldsymbol{T}_{N}^{p})\}\), corresponding to left eye, right eye, and patient level, respectively. The image and text features of the same patient in each subset are positive samples of each other, while the rest are negative samples. The cosine similarity matrix is calculated on each subset.

For the subset of left eye features, we obtain the image feature matrix \(\mathbf{V}^{l}=(\boldsymbol{V}_{1}^{l},\cdots,\boldsymbol{V}_{N}^{l})\in\mathbb{R}^{N \times d}\) and the text feature matrix \(\mathbf{T}^{l}=(\boldsymbol{T}_{1}^{l},\cdots,\boldsymbol{T}_{N}^{l})\in\mathbb{R}^{N \times d}\). We measure the inter-sample similarity, termed as \(\mathbf{P}^{v2t}\) and \(\mathbf{P}^{t2v}\), using the cosine distance \(S(\cdot)\):
\begin{equation}
    \mathbf{P}^{v2t}=S(\mathbf{V}^{l},\mathbf{T}^{l})\in\mathbb{R}^{N \times N}, \mathrm{~}\mathbf{P}^{t2v}=S(\mathbf{T}^{l},\mathbf{V}^{l})\in\mathbb{R}^{N \times N}.
    \label{eq6}
\end{equation}
Then we calculate the contrastive loss of the left eye:
\begin{equation}
    \mathcal{L}_{l}=\frac{1}{2}(\mathrm{CE}(\mathbf{P}^{v2t},\mathbf{Y}^{v2t})+\mathrm{CE}(\mathbf{P}^{t2v},\mathbf{Y}^{t2v}),
    \label{eq7}
\end{equation}
where \(\mathbf{Y}^{v2t}\) and \(\mathbf{Y}^{t2v}\) represent the one-hot labels, \(\mathrm{CE}\) refers to InfoNCE loss \cite{oord2018representation}.

Then we calculate \(\mathcal{L}_{r}\) and \(\mathcal{L}_{p}\) for right eye and patient level based on \(\mathcal{F}^{r}\) and \(\mathcal{F}^{p}\) in the same way. The final loss is the sum of the above three:
\begin{equation}
    \mathcal{L}=\mathcal{L}_{l}+\mathcal{L}_{r}+\mathcal{L}_{p}.
    \label{eq8}
\end{equation}

\subsection{Implementation}
The vision encoder utilizes the base-sized version of the vision transformer (ViT-base) \cite{dosovitskiy2020image}, while the text encoder employs the base-sized version of RoBERTa (RoBERTa-base) \cite{liu2019roberta}, both are initialized with the Chinese-CLIP weights \cite{yang2022chinese}.
AdamW is used as the optimizer. The batch size is 256, and training is performed using NVIDIA GeForce RTX 4090. The training process consists of 10 epochs, with the first 50 steps dedicated to warming up the model (from 0 to a learning rate of \(3\times10^{-5}\)).

\section{Experiments}
\subsection{Tasks and Datasets}
We focus on designing downstream evaluation experiments primarily for visual tasks. These tasks contain four main categories: diagnosis of diabetic retinopathy, glaucoma, multiple diseases, and multi-label classification of multiple diseases.

\textbf{For diabetic retinopathy diagnosis}, IDRID \cite{porwal2020idrid} and APTOS-2019 (https://www.kaggle.com/competitions/aptos2019-blindness-detection/data) are used. The labels for diabetic retinopathy are no, mild, moderate, severe, and proliferative retinopathy. The IDRID dataset comprises 516 images, while the APTOS dataset contains 3662 images.

\textbf{For glaucoma diagnosis}, PAPILA \cite{kovalyk2022papila} (488 images in total) and Glaucoma Fundus \cite{ahn2018deep} (1544 images in total) are used. They both have three categorical labels, non-glaucoma, suspected glaucoma (early glaucoma), and glaucoma (advanced glaucoma). 

\textbf{For multiple disease diagnosis}, JSIEC \cite{cen2021automatic} (1000 in total) and Retina (https://www.kaggle.com/datasets/jr2ngb/cataractdataset) (601 in total) are tested. JSIEC contains 39 categories of common referable fundus diseases and conditions. Retina includes labels for normal, glaucoma, cataract, and other retinal diseases.

\textbf{For multi-label classification of multiple diseases}, RFMID \cite{pachade2021retinal} and ODIR (https://odir2019.grand-challenge.org/) are tested. RFMID includes 3200 images with 28 categories of common referable fundus diseases and conditions. ODIR includes 10000 images (5000 patients' paired left and right eyes) with labels of normal, diabetic retinopathy, glaucoma, cataract,age-related macular degeneration (AMD), hypertension, myopia, and other diseases.

For the IDRIR, the entire dataset is officially divided into a test set comprising 20\% of the data, with the remaining 80\% designated as the training set. In our experiments, we further split the training set into a training set and a validation set using a 4:1 ratio. Similarly, for the PAPLA, we follow the official partitioning method, which aligns with the approach described above. Regarding the RFMID, the official division includes distinct sets for training, validation, and testing; we adhere to this official partitioning. For all other datasets, we divide them into training, validation, and test sets using a 0.56:0.14:0.3 ratio, following RETFound's \cite{zhou2023foundation} partitioning method. For all datasets, samples within each category are initially distributed based on the specified proportions before being combined to ensure consistent category distribution across the training, validation, and test sets.

When adapting to downstream tasks, the input image is mapped to a high-level feature representation by the visual encoder. A simple linear prediction head is then applied, followed by a Sigmoid or Softmax layer to achieve classification. 

For each task, two adaptation methods are implemented: linear probing, training the classifier only with the encoder frozen, and fine-tuning, where both the encoder and classifier are trained. Each evaluation process consists of 50 epochs with a batch size of 16. The model weights with the best performance on the validation set are saved for testing.
\subsection{Comparision Methods and Evaluation Metrics}
To demonstrate the superiority of our method, we compare two broad categories of models: foundation models trained on non-CFP datasets (Chinese-CLIP \cite{yang2022chinese}, PMC-CLIP \cite{lin2023pmc}, DINOv2 \cite{oquab2023dinov2}) and models designed for CFP vision tasks (RETFound \cite{zhou2023foundation}, FLAIR \cite{silva2023foundation}).

We use the area under the receiver operating curve (AUROC) and area under the precision-recall curve (AUPR) as the evaluation metrics. We evaluate five iterations with different random seeds for each model on each downstream dataset to calculate the mean values. We also conduct the t-test for each downstream task to determine the significance level at which the top-performing method surpasses the others (see Supplementary Materials).

\subsection{Result}
RET-CLIP outperforms five comparison models across eight datasets (four categories) as introduced before, demonstrating strong generalization capabilities.

\textbf{For linear probing}, the results are shown in \Cref{tab1} and \Cref{tab2}. RET-CLIP demonstrates superior performance on almost all datasets, which indicates that RET-CLIP has learned a rich feature representation during the pre-training phase, demonstrating the capability to capture high-quality features.

\begin{table}[]
\centering
\caption{Diabetic retinopathy and glaucoma diagnosis results for linear probing. The best results on each metric are highlighted in bold.}
\label{tab1}
\resizebox{\textwidth}{!}{%
\begin{tabular}{c|cc|cc|cc|cc}
\toprule
\multirow{2}{*}{\textbf{Models}} & \multicolumn{2}{c|}{\textbf{IDRID}} & \multicolumn{2}{c|}{\textbf{APTOS2019}} & \multicolumn{2}{c|}{\textbf{PAPILA}} & \multicolumn{2}{c}{\textbf{Glaucoma Fundus}} \\
 & \textbf{AUROC} & \textbf{AUPR} & \textbf{AUROC} & \textbf{AUPR} & \textbf{AUROC} & \textbf{AUPR} & \textbf{AUROC} & \textbf{AUPR} \\ \hline
\textbf{CN-CLIP \cite{yang2022chinese}} & 0.633 & 0.336 & 0.806 & 0.429 & 0.658 & 0.473 & 0.863 & 0.716 \\
\textbf{PMC-CLIP \cite{lin2023pmc}} & 0.585 & 0.303 & 0.756 & 0.368 & 0.773 & 0.603 & 0.899 & 0.780 \\
\textbf{DinoV2 \cite{oquab2023dinov2}} & 0.748 & 0.463 & 0.783 & 0.432 & 0.740 & 0.556 & 0.891 & 0.746 \\
\textbf{RETFound \cite{zhou2023foundation}} & 0.665 & 0.368 & 0.745 & 0.370 & 0.620 & 0.511 & \textbf{0.899} & 0.773 \\
\textbf{FLAIR \cite{silva2023foundation}} & 0.700 & 0.475 & 0.849 & 0.515 & 0.746 & 0.595 & 0.872 & 0.672 \\ \hline
\textbf{OURS} & \textbf{0.856} & \textbf{0.616} & \textbf{0.923} & \textbf{0.656} & \textbf{0.775} & \textbf{0.667} & 0.893 & \textbf{0.789} \\ \bottomrule
\end{tabular}%
}
\end{table}
\begin{table}[]
\centering
\caption{Multiple disease diagnosis and multi-label classification of multiple diseases results for linear probing.}
\label{tab2}
\resizebox{\textwidth}{!}{%
\begin{tabular}{c|cc|cc|cc|cc}
\toprule
\multirow{2}{*}{\textbf{Models}} & \multicolumn{2}{c|}{\textbf{JSIEC}} & \multicolumn{2}{c|}{\textbf{Retina}} & \multicolumn{2}{c|}{\textbf{RFMID}} & \multicolumn{2}{c}{\textbf{ODIR}} \\
 & \textbf{AUROC} & \textbf{AUPR} & \textbf{AUROC} & \textbf{AUPR} & \textbf{AUROC} & \textbf{AUPR} & \textbf{AUROC} & \textbf{AUPR} \\ \hline
\textbf{CN-CLIP \cite{yang2022chinese}} & 0.783 & 0.239 & 0.738 & 0.514 & 0.819 & 0.293 & 0.801 & 0.483 \\
\textbf{PMC-CLIP \cite{lin2023pmc}} & 0.947 & 0.654 & 0.778 & 0.597 & 0.854 & 0.372 & 0.800 & 0.506 \\
\textbf{DinoV2 \cite{oquab2023dinov2}} & 0.873 & 0.446 & 0.813 & 0.635 & 0.860 & 0.430 & 0.825 & 0.550 \\
\textbf{RETFound \cite{zhou2023foundation}} & 0.704 & 0.167 & 0.630 & 0.434 & 0.842 & 0.409 & 0.738 & 0.401 \\
\textbf{FLAIR \cite{silva2023foundation}} & 0.843 & 0.304 & 0.773 & 0.557 & 0.773 & 0.254 & 0.858 & 0.531 \\ \hline
\textbf{OURS} & \textbf{0.982} & \textbf{0.855} & \textbf{0.935} & \textbf{0.864} & \textbf{0.925} & \textbf{0.552} & \textbf{0.902} & \textbf{0.682} \\ \bottomrule
\end{tabular}%
}
\end{table}
\textbf{For fine-tuning}, as shown in \Cref{tab3} and \Cref{tab4}, RET-CLIP demonstrates superior performance across nearly all tasks. This outcome substantiates RET-CLIP's robust feature extraction and generalization capabilities. Furthermore, 
it suggests that RET-CLIP not only captures high-quality features but also exhibits strong adaptability, enabling effective customization for specific tasks.

\begin{table}[]
\centering
\caption{Diabetic retinopathy and glaucoma diagnosis results for fine-tuning.}
\label{tab3}
\resizebox{\textwidth}{!}{%
\begin{tabular}{c|cc|cc|cc|cc}
\toprule
\multirow{2}{*}{\textbf{Models}} & \multicolumn{2}{c|}{\textbf{IDRID}} & \multicolumn{2}{c|}{\textbf{APTOS2019}} & \multicolumn{2}{c|}{\textbf{PAPILA}} & \multicolumn{2}{c}{\textbf{Glaucoma Fundus}} \\
 & \textbf{AUROC} & \textbf{AUPR} & \textbf{AUROC} & \textbf{AUPR} & \textbf{AUROC} & \textbf{AUPR} & \textbf{AUROC} & \textbf{AUPR} \\ \hline
\textbf{CN-CLIP \cite{yang2022chinese}} & 0.778 & 0.506 & 0.881 & 0.619 & 0.804 & 0.690 & 0.951 & 0.876 \\
\textbf{PMC-CLIP \cite{lin2023pmc}} & 0.785 & 0.511 & 0.776 & 0.386 & 0.798 & 0.659 & 0.925 & 0.827 \\
\textbf{DinoV2 \cite{oquab2023dinov2}} & 0.791 & 0.533 & 0.920 & 0.675 & 0.797 & 0.681 & 0.955 & 0.884 \\
\textbf{RETFound \cite{zhou2023foundation}} & 0.822 & 0.496 & 0.943 & 0.726 & \textbf{0.855} & 0.748 & 0.943 & 0.863 \\
\textbf{FLAIR \cite{silva2023foundation}} & 0.795 & 0.529 & 0.932 & 0.686 & 0.752 & 0.610 & 0.905 & 0.792 \\ \hline
\textbf{OURS} & \textbf{0.863} & \textbf{0.630} & \textbf{0.951} & \textbf{0.748} & 0.853 & \textbf{0.754} & \textbf{0.958} & \textbf{0.889} \\ \bottomrule
\end{tabular}%
}
\end{table}
\begin{table}[]
\centering
\caption{Multiple disease diagnosis and multi-label classification of multiple diseases results for fine-tuning.}
\label{tab4}
\resizebox{\textwidth}{!}{%
\begin{tabular}{c|cc|cc|cc|cc}
\toprule
\multirow{2}{*}{\textbf{Models}} & \multicolumn{2}{c|}{\textbf{JSIEC}} & \multicolumn{2}{c|}{\textbf{Retina}} & \multicolumn{2}{c|}{\textbf{RFMID}} & \multicolumn{2}{c}{\textbf{ODIR}} \\
 & \textbf{AUROC} & \textbf{AUPR} & \textbf{AUROC} & \textbf{AUPR} & \textbf{AUROC} & \textbf{AUPR} & \textbf{AUROC} & \textbf{AUPR} \\ \hline
\textbf{CN-CLIP \cite{yang2022chinese}} & 0.992 & 0.882 & 0.839 & 0.691 & 0.901 & 0.480 & 0.859 & 0.598 \\
\textbf{PMC-CLIP \cite{lin2023pmc}} & 0.964 & 0.738 & 0.875 & 0.742 & 0.894 & 0.456 & 0.819 & 0.542 \\
\textbf{DinoV2 \cite{oquab2023dinov2}} & 0.996 & 0.918 & 0.893 & 0.771 & 0.914 & 0.547 & 0.867 & 0.621 \\
\textbf{RETFound \cite{zhou2023foundation}} & 0.990 & 0.884 & 0.847 & 0.697 & 0.889 & 0.489 & 0.850 & 0.620 \\
\textbf{FLAIR \cite{silva2023foundation}} & 0.917 & 0.704 & 0.863 & 0.679 & 0.870 & 0.397 & 0.860 & 0.601 \\ \hline
\textbf{OURS} & \textbf{0.999} & \textbf{0.972} & \textbf{0.942} & \textbf{0.871} & \textbf{0.946} & \textbf{0.581} & \textbf{0.917} & \textbf{0.715} \\ \bottomrule
\end{tabular}%
}
\end{table}
It's noteworthy that the previous foundation models designed for CFPs do not exhibit an advantage over models trained on non-CFP datasets. RETFound's \cite{zhou2023foundation} image reconstruction-focused paradigm may prioritize features related to the rebuilding of CFP, which lack the granularity and quality needed for specific downstream tasks, hindering its broader applicability. FLAIR \cite{silva2023foundation}, while is a CLIP-style model, does not suit ophthalmic tasks as it uses the text provision method employed by the original CLIP \cite{radford2021learning}, which is designed for natural contexts, offering limited textual insights from single labels. Moreover, its dependence on public datasets for training constrains its performance due to their limited scale and quality. In contrast, RET-CLIP leverages rich textual information from clinical reports to extract detailed features for ophthalmic tasks better, showcasing the benefits of integrating diagnostic reports into the training of medical CLIP-style models.
\subsection{Ablation study}
The results, as shown in \Cref{tab5}, confirm the effectiveness of optimizing objectives at both monocular and patient levels. As previously discussed, the combination of the global information provided at the patient level with the finer-grained features contributed at the monocular level is essential to achieve optimal performance.
\begin{table}[]
\centering
\caption{Results of ablation studies. Monocular-level loss refers to \(\mathcal{L}_{l}\) plus \(\mathcal{L}_{r}\).}
\label{tab5}
\begin{tabularx}{\textwidth}{CCCCCCC}
\toprule
\multicolumn{1}{c|}{} & \multicolumn{3}{c|}{\textbf{AUROC}} & \multicolumn{3}{c}{\textbf{AUPR}} \\
\multicolumn{1}{c|}{\textbf{Monocular-level Loss}} & \checkmark & \textbf{} & \multicolumn{1}{C|}{\checkmark} & \checkmark & \textbf{} & \multicolumn{1}{c}{\checkmark} \\
\multicolumn{1}{c|}{\textbf{Patient-level Loss}} & \checkmark & \checkmark & \multicolumn{1}{C|}{\textbf{}} & \checkmark & \checkmark & \multicolumn{1}{c}{\textbf{}} \\ \hline
\multicolumn{1}{c|}{\textbf{IDRID}}                & \textbf{0.863} & 0.860           & \multicolumn{1}{C|}{0.847}      & \textbf{0.63}  & 0.623      & 0.619          \\
\multicolumn{1}{c|}{\textbf{APTOS-2019}}    & \textbf{0.951} & 0.945          & \multicolumn{1}{C|}{0.941}      & 0.748          & 0.737      & \textbf{0.759} \\
\multicolumn{1}{c|}{\textbf{PAPILA}}               & 0.853          & \textbf{0.864} & \multicolumn{1}{C|}{0.846}      & \textbf{0.754} & 0.745      & 0.739          \\
\multicolumn{1}{c|}{\textbf{Glaucoma Fundus}}      & \textbf{0.958} & 0.948          & \multicolumn{1}{C|}{0.957}      & \textbf{0.889} & 0.869      & 0.888          \\
\multicolumn{1}{c|}{\textbf{JSIEC}}                & \textbf{0.999} & 0.997          & \multicolumn{1}{C|}{0.997}      & \textbf{0.972} & 0.949      & 0.962          \\
\multicolumn{1}{c|}{\textbf{Retina}}               & \textbf{0.942} & 0.939          & \multicolumn{1}{C|}{0.935}      & 0.871          & 0.869      & \textbf{0.876} \\
\multicolumn{1}{c|}{\textbf{RFMID}}                & \textbf{0.946} & 0.924          & \multicolumn{1}{C|}{0.940}       & \textbf{0.581} & 0.573      & 0.578          \\
\multicolumn{1}{c|}{\textbf{ODIR}}                 & \textbf{0.917} & 0.909          & \multicolumn{1}{C|}{0.905}      & \textbf{0.715} & 0.692      & 0.696  \\ \bottomrule
\end{tabularx}%
\end{table}
\section{Conclusion}
In this study, we compile a binocular images-text dataset, RET-Clinical, derived from 193,865 clinical patients, with which, we jointly optimize and pre-train a CLIP-style model, RET-CLIP, cooperating with the information of left eye, right eye, and patient level. RET-CLIP achieves state-of-the-art results across eight downstream tasks spanning four critical diagnostic categories.  Our research narrows the existing void in ophthalmic vision-language models by integrating textual data from clinical diagnostic reports, thereby offering insights into the applicability of raw clinical texts in the wider medical domain.




\begin{credits}
\subsubsection{\ackname} This research work is supported by 
Beijing Natural Science Foundation (Grant No. IS23112), National Natural
Science Foundation of China (NSFC) (GrantNo. 82072007), and Beijing Institute of Technology Research Fund Program for Young Scholars.

\subsubsection{\discintname}
The authors have no competing interests to declare that are
relevant to the content of this article. 
\end{credits}
%
%
%
\newpage
\bibliographystyle{splncs04}
\bibliography{Paper}
%




\end{sloppypar}
\end{document}